\documentclass[sigconf]{acmart}

\usepackage{enumitem}                      

\usepackage{xspace}

\newcommand{\sysname}{\textit{AdMuse}\xspace}

\AtBeginDocument{%
  }

\copyrightyear{2026}
\acmYear{2026}
\setcopyright{cc}
\setcctype{by}
\acmConference[SA Technical Communications '26]{SIGGRAPH Asia 2026 Technical Communications}{December 01--04, 2026}{Kuala Lumpur, Malaysia}
\acmBooktitle{SIGGRAPH Asia 2026 Technical Communications (SA Technical Communications '26), December 01--04, 2026, Kuala Lumpur, Malaysia}
\acmDOI{10.1145/3829339.3847841}
\acmISBN{979-8-4007-2841-9/2026/12}

\begin{document}

\title{AdMuse: Skill-Guided Creative Advertising Image Generation}


\author{Minyan Luo}
\orcid{0009-0008-2675-1650}
\affiliation{%
  \institution{MAIS, Institute of Automation, CAS}
  \city{Beijing}
  \country{China}
}
\affiliation{%
  \institution{School of Artificial Intelligence, UCAS}
  \city{Beijing}
  \country{China}
}
\email{luominyan2025@ia.ac.cn}

\author{Yuxin Zhang}
\affiliation{%
  \institution{MAIS, Institute of Automation, CAS}
  \city{Beijing}
  \country{China}
}
\affiliation{%
  \institution{School of Artificial Intelligence, UCAS}
  \city{Beijing}
  \country{China}
}
\email{yuxin.zhazel@gmail.com}

\author{Meng Zhao}
\affiliation{%
  \institution{School of Artificial Intelligence, UCAS}
  \city{Beijing}
  \country{China}
}
\email{zhaomeng24@mails.ucas.ac.cn}

\author{Xincan Wang}
\affiliation{%
  \institution{Shanghai Theatre Academy}
  \city{Shanghai}
  \country{China}
}
\email{wangxincan@sta.edu.cn}

\author{Fuzhang Wu}
\affiliation{%
  \institution{Institute of Software, CAS}
  \city{Beijing}
  \country{China}
}
\email{fuzhang@iscas.ac.cn}

\author{Tong-Yee Lee}
\affiliation{%
  \institution{National Cheng Kung University}
  \city{Tainan}
  \country{Taiwan}
}
\email{tonylee@mail.ncku.edu.tw}

\author{Oliver Deussen}
\affiliation{%
  \institution{University of Konstanz}
  \city{Konstanz}
  \country{Germany}
}
\email{oliver.deussen@uni-konstanz.de}

\author{Fan Tang}
\affiliation{%
  \institution{ICT, CAS}
  \city{Beijing}
  \country{China}
}
\email{tfan.108@gmail.com}

\author{Weiming Dong}
\authornote{Corresponding author}
\affiliation{%
  \institution{MAIS, Institute of Automation, CAS}
  \city{Beijing}
  \country{China}
}
\email{weiming.dong@ia.ac.cn}

\renewcommand{\shortauthors}{Luo et al.}

\begin{abstract}
An outstanding creative advertising image should both attract attention and communicate a product's positioning.
How to choose a persuasive conceptual bridge before rendering begins is challenging for advertising generation.
We present \sysname, a skill-guided multi-agent framework that distills high-quality advertisements into reusable creative skills for cross-domain advertising strategy. Rather than retrieving isolated exemplars, \sysname abstracts recurring bridge patterns into a skill library that can guide new product briefs. At generation time, \sysname coordinates skill routing, multi-candidate ideation and selection, and VLM-guided visual refinement: diverse skills propose alternative concepts, relative judgment selects the most promising direction, and vision-language feedback helps realize the selected intent in the final image. This design treats advertising creativity as a structured decision process rather than a single prompt-writing step. Experiments across diverse product and brand positionings show that \sysname produces visually compelling advertisements that faithfully communicate product positioning.
\end{abstract}

\begin{teaserfigure}
\centering
\includegraphics[width=0.82\linewidth]{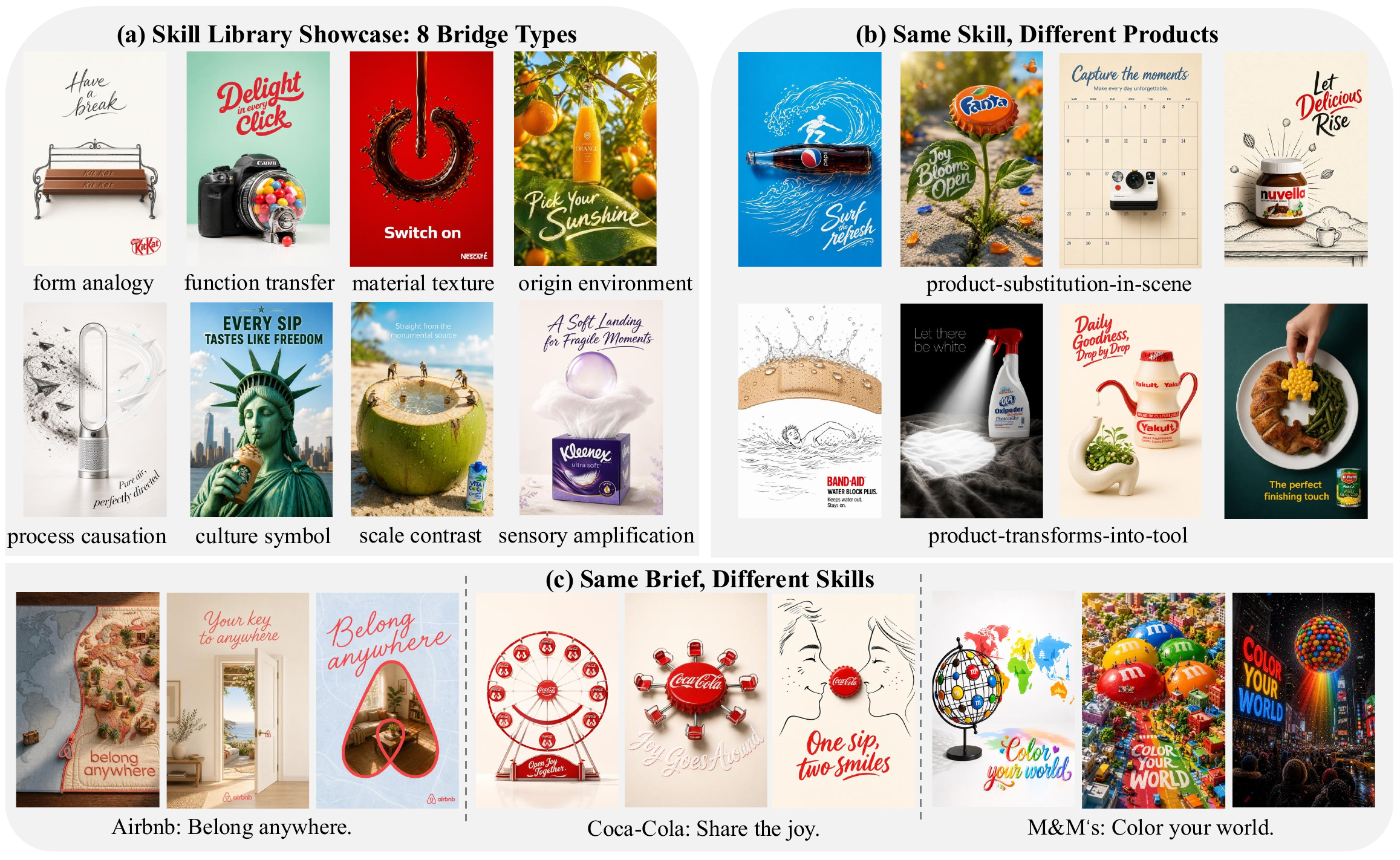}
\caption{%
    \textbf{Creative advertising images generated by AdMuse.}
    \textbf{(a)} One representative ad per bridge type from skill library.
    \textbf{(b)} A single skill applied across diverse products.
    \textbf{(c)} A single brief routed through different skills yields visually distinct ads.%
  }
\label{fig:teaser}
\end{teaserfigure}

\maketitle

\begin{sloppypar}


\begin{figure*}
\centering
\includegraphics[width=0.85\linewidth]{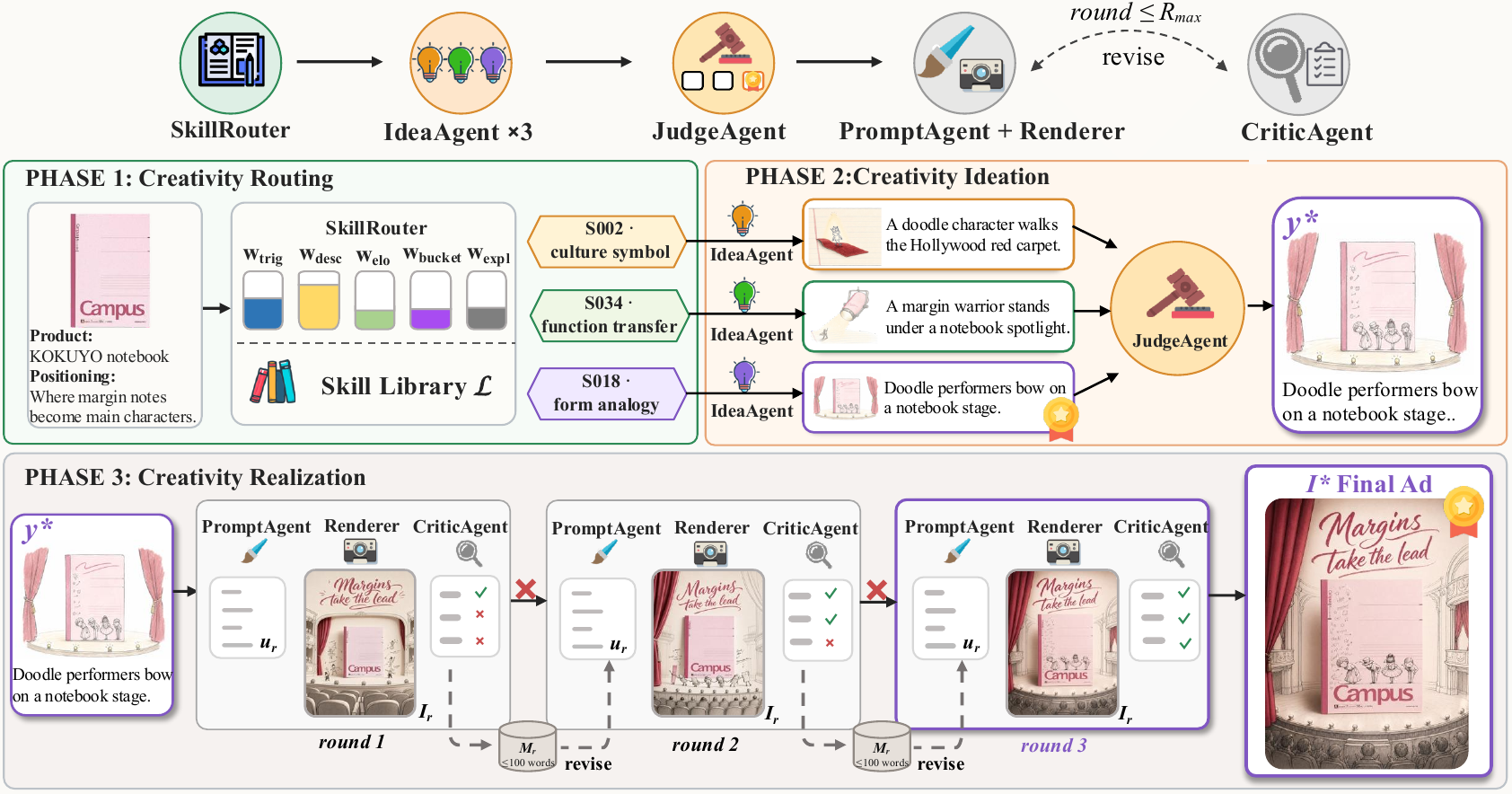}
\caption{
\sysname combines a creative skill library with a three-phase generation pipeline.
}
\label{fig_pipeline}
\end{figure*}

\section{Introduction}
\label{sec:intro}

Creativity plays a vital role in advertising. Creative advertisements
are more memorable than regular ones~\cite{till2005recall,jin2019impairment}. Although modern image generators~\cite{openai2026gptimage2} and agentic visual pipelines~\cite{chen2026agentic} can render high-quality images, they still largely rely on the user to provide the core creative idea. For advertising, however, the main difficulty often lies not in rendering an image, but in deciding \emph{what visual concept to use}.
A key challenge is that a product can support many creative directions, each producing distinct visual and communicative effects. Advertising generation therefore needs a creative-strategy representation that reveals, compares, and instantiates viable directions~\cite{koushik2025mind}. It should generalize across products while guiding visual synthesis, often through cross-domain associations that make desired qualities perceptible.

We frame this problem through the notion of a \emph{creative bridge}: a cross-domain association that links a product to a distant concept through an interpretable visual chain. High-quality creative ads repeatedly reuse a limited set of such bridge patterns~\cite{sun2025creative}. We therefore distill them into reusable \emph{creative skills} that specify when a strategy applies, what associative chain it constructs, and how it can be instantiated for a new product brief.
Based on this representation, we propose \textbf{\sysname}, a skill-guided multi-agent framework for creative advertising image generation. \sysname routes bridge-diverse skills for a brief, lets multiple skill-conditioned concepts compete through relative judgment, and then refines the selected concept with VLM feedback. 
Our contributions are summarized as:
\begin{itemize}[nosep, leftmargin=*]
  \item We formalize the \emph{creative bridge} and distill recurring bridge patterns from advertisements into a reusable skill library.
  \item We propose a skill-guided multi-agent pipeline that combines routing, multi-candidate ideation and selection, and VLM-guided visual refinement.
  \item We show that \sysname produces visually compelling and message-faithful advertisements across diverse briefs, products, and positionings (see Fig.~\ref{fig:teaser}).
\end{itemize}


\section{Related Work}
\label{sec:related}

Recent agentic image-generation systems improve synthesis through multi-agent coordination~\cite{shen2026agentic} and VLM-guided critique~\cite{yaophotoagent}. In parallel, skill- and memory-based agents show how reusable procedures can guide complex tasks~\cite{zhong2026skilllearnbench}. Our work connects these directions by treating advertising ideation itself as a reusable, skill-based strategy before rendering.
Creative advertising also relates to visual metaphor and conceptual blending~\cite{sun2025creative,koushik2025mind}, where a product is linked to a remote concept through a meaningful visual association. Prior methods typically generate one metaphor per query or rely on direct reference transfer~\cite{xu2026beyond}. By contrast, we distill recurring bridge patterns from high-quality ads into reusable skills that can be routed and compared for a new brief.
In e-commerce and marketing AIGC, existing systems focus on product staging~\cite{wang2025generate}, poster composition~\cite{gao2025postermaker}, or marketing copy generation~\cite{quan2025crmagent}. These approaches usually assume the core idea is already given. We instead address the upstream problem of selecting a persuasive visual concept that communicates product positioning.
\section{Method}
\label{sec:method}

\subsection{Creative Bridges and Skill Distillation}
\label{sec:method-bridge-skill}

The overall framework of \sysname is shown in Fig.~\ref{fig_pipeline}.
We build the skill library from $1{,}129$ creative advertising images spanning 13 product categories. We use Gemini-3.1-Pro to analyze each advertisement and extract a structured record of its product identity, associative space, creative bridge, visual structure, and creative reasoning. Details of the corpus and distillation procedure are provided in Supplementary Sec.~1.

Creative advertising often expresses an abstract product attribute by connecting the product to a remote visual domain, a process closely related to conceptual blending theory~\cite{fauconnier2008way}.
We operationalize this mechanism as a \emph{creative bridge}: a cross-domain mapping from the product source to the target attribute specified by the positioning.
A bridge records the intermediate concepts that carry the mapping, together with the order in which they are visually interpreted.
Formally,
\begin{equation}
  bridge:\quad \mathrm{product\ source} \;\xrightarrow[\;bridge\ type\;]{\;n_1 \,\to\, n_2 \,\to\, \cdots \,\to\, n_K\;}\; \mathrm{target},
  \label{eq:bridge}
\end{equation}
where $\mathrm{target}$ denotes the attribute to be communicated, $n_1,\ldots,n_K$ ($K\geq 1$) denotes an ordered chain of source-domain concepts, and the bridge type specifies the relation that makes the mapping plausible.

We derive the bridge taxonomy from the annotated corpus rather than prescribing categories manually: recurring mapping relations are consolidated into eight bridge types, after which every advertisement is re-labeled with the taxonomy fixed and assigned to exactly one type:
$\mathcal{B} = \{$\texttt{form\_analogy}, \texttt{function\_transfer}, \texttt{material\_texture}, \texttt{origin\_environment}, \texttt{cultural\_symbol}, \texttt{scale\_contrast}, \texttt{process\_causation}, \texttt{sensory\_amplification}$\}$.

Bridge types provide a coarse organization of creative relations, but they are not sufficiently specific to guide generation directly.
We therefore cluster normalized embeddings of each record's associative chain, rationale, and creative reasoning within each bridge family using K-Means, and summarize each cluster as a reusable \emph{skill}.
A skill packages a reusable creative procedure: a strategy description, bridge type, applicability triggers and anti-triggers, a parameterized associative-chain template, and a playbook specifying construction steps, visual structure, constraints, anti-patterns, and canonical examples. 
The complete schema and the S018 skill are provided in Supplementary Sec. 1.3.

\subsection{Skill-Guided Generation Pipeline}
\label{sec:method-pipeline}

At inference, \sysname maps a brief $x=(n,z,\mathbf{I}_0)$, i.e., product name, positioning statement, and optional product image, to a creative advertising image $\mathbf{I}^*$ through routing, ideation, and realization:
\begin{equation}
\mathcal{S}_x = \mathcal{F}_{\mathrm{route}}(x,\mathcal{L}), \quad
y^* = \mathcal{F}_{\mathrm{idea}}(x,\mathcal{S}_x), \quad
\mathbf{I}^* = \mathcal{F}_{\mathrm{realize}}(y^*,x),
\label{eq:pipeline}
\end{equation}
where $\mathcal{L}$ is the skill library, $\mathcal{S}_x$ denotes the selected skills, and $y^*$ denotes the winning creative concept (see Fig.~\ref{fig_pipeline}).

\paragraph{Phase 1: Creativity Routing}
For the input brief $x=(n,z,\mathbf{I}_0)$, the \textsc{SkillRouter} considers trigger match ($w_{\mathrm{trig}}$), description similarity ($w_{\mathrm{desc}}$), global skill evidence ($w_{\mathrm{Elo}}$), category-specific evidence ($w_{\mathrm{bucket}}$), and an exploration bonus ($w_{\mathrm{expl}}$).
It greedily selects three skills with distinct bridge types to expose the subsequent ideation phase to different creative mechanisms. The scoring and selection details are provided in Supplementary Sec.~2.1.

\paragraph{Phase 2: Creativity Ideation}
For each selected skill $s_i$, the \textsc{IdeaAgent} $\mathcal{G}_{\mathrm{idea}}$ generates a structured concept $y_i$ conditioned on the shared brief $x$.
Each concept specifies the metaphor core, visual composition, and tagline.
The \textsc{JudgeAgent} $\mathcal{G}_{\mathrm{judge}}$ then compares the three concepts and selects the winning idea,
\begin{equation}
  y_i = \mathcal{G}_{\mathrm{idea}}(x,s_i), \qquad
  y^* = \mathcal{G}_{\mathrm{judge}}(y_1,y_2,y_3),
  \label{eq:creativity-ideation}
\end{equation}
using positioning alignment, creative novelty, and visual executability as criteria.
Since no image is rendered in this phase, \sysname separates strategic concept selection from visual realization.

\paragraph{Phase 3: Creativity Realization}
The \textsc{PromptAgent} converts the winning concept $y^*$ into an image prompt, which the renderer uses to generate an advertising image.
The \textsc{CriticAgent} inspects the rendered image against the selected concept and either approves it or provides feedback for prompt revision.
Rejected attempts are summarized into a compact memory to prevent repeated failures, and the loop terminates upon approval or after $R_{\max}$ rounds.


\section{Experiments}
\label{sec:exp}

\subsection{Experimental Setup}
\label{sec:exp-setup}

\paragraph{Implementation Details}
Our pipeline uses GPT-5.5 for the \textsc{IdeaAgent} and \textsc{PromptAgent}, Gemini-3.1-Pro for in-loop judging and critique, and GPT-Image-2 for image synthesis. Claude Sonnet 4.6 and Qwen3.7-Plus independently evaluate the same final outputs using a shared prompt and rubrics, reducing dependence on a single VLM evaluator~\cite{li2025preference}. Further implementation details and hyperparameters are provided in Supplementary Sec.~2.3--2.4.

\paragraph{Benchmark and Baselines}
We evaluate on 500 briefs covering 420 product images across 13 categories, held out from the 1,129-ad distillation corpus to prevent leakage. We compare \sysname with four baselines: \textbf{Direct} sends the brief directly to the image renderer (GPT-Image-2); \textbf{LLM-Rewrite} expands the brief into a detailed image-generation prompt; \textbf{RAG-Retrieve} retrieves one visual sketch from the reference corpus as inspiration; and \textbf{RAG-Refine},  feeds the top-$3$ retrieved ad records into the same idea--judge--prompt--critic loop as our system, but without skill routing or templates. Detailed implementations are provided in Supplementary Sec.~3.2.

\subsection{Quantitative Evaluation}
\label{sec:exp-quant}

We report two evaluation metrics: \textbf{Creativity Score (CS)}, which measures the novelty and conceptual impact of the advertisement, and \textbf{Alignment Score (AS)}, which measures how clearly the image communicates the intended positioning while preserving product identity. Both scores are assigned by two independent out-of-loop VLM judges under the same protocol. Full rubrics and prompt details are deferred to Supplementary Sec.~3.3. As shown in Table~\ref{tab:main-results}, \sysname achieves the highest average, CS, and AS scores on the 500-brief benchmark, outperforming all four baselines.

\begin{table}[t]
\centering
\caption{Quantitative evaluation results.
}
\label{tab:main-results}
\small
\setlength{\tabcolsep}{3pt}
\begin{tabular}{lcccccc}
\toprule
& \multicolumn{3}{c}{Claude Sonnet 4.6} & \multicolumn{3}{c}{Qwen3.7-Plus} \\
\cmidrule(lr){2-4}\cmidrule(lr){5-7}
Method & CS$\uparrow$ & AS$\uparrow$ & Avg$\uparrow$ & CS$\uparrow$ & AS$\uparrow$ & Avg$\uparrow$ \\
\midrule
Direct       & 5.046 & 6.168 & 5.607 & 5.115 & 6.916 & 6.016 \\
LLM-Rewrite  & 5.623 & 6.782 & 6.202 & 5.743 & 7.566 & 6.655 \\
RAG-Retrieve & 6.633 & 7.262 & 6.948 & 6.704 & 7.752 & 7.228 \\
RAG-Refine & 7.192 & 7.317 & 7.255 & 6.619 & 6.902 & 6.760 \\
\midrule
AdMuse       & \textbf{7.295} & \textbf{7.618} & \textbf{7.457} & \textbf{7.482} & \textbf{8.232} & \textbf{7.857} \\
\bottomrule
\end{tabular}
\end{table}

\subsection{Qualitative Evaluation}
\label{sec:exp-qual}

\begin{figure}
  \centering

\includegraphics[width=0.95\linewidth]{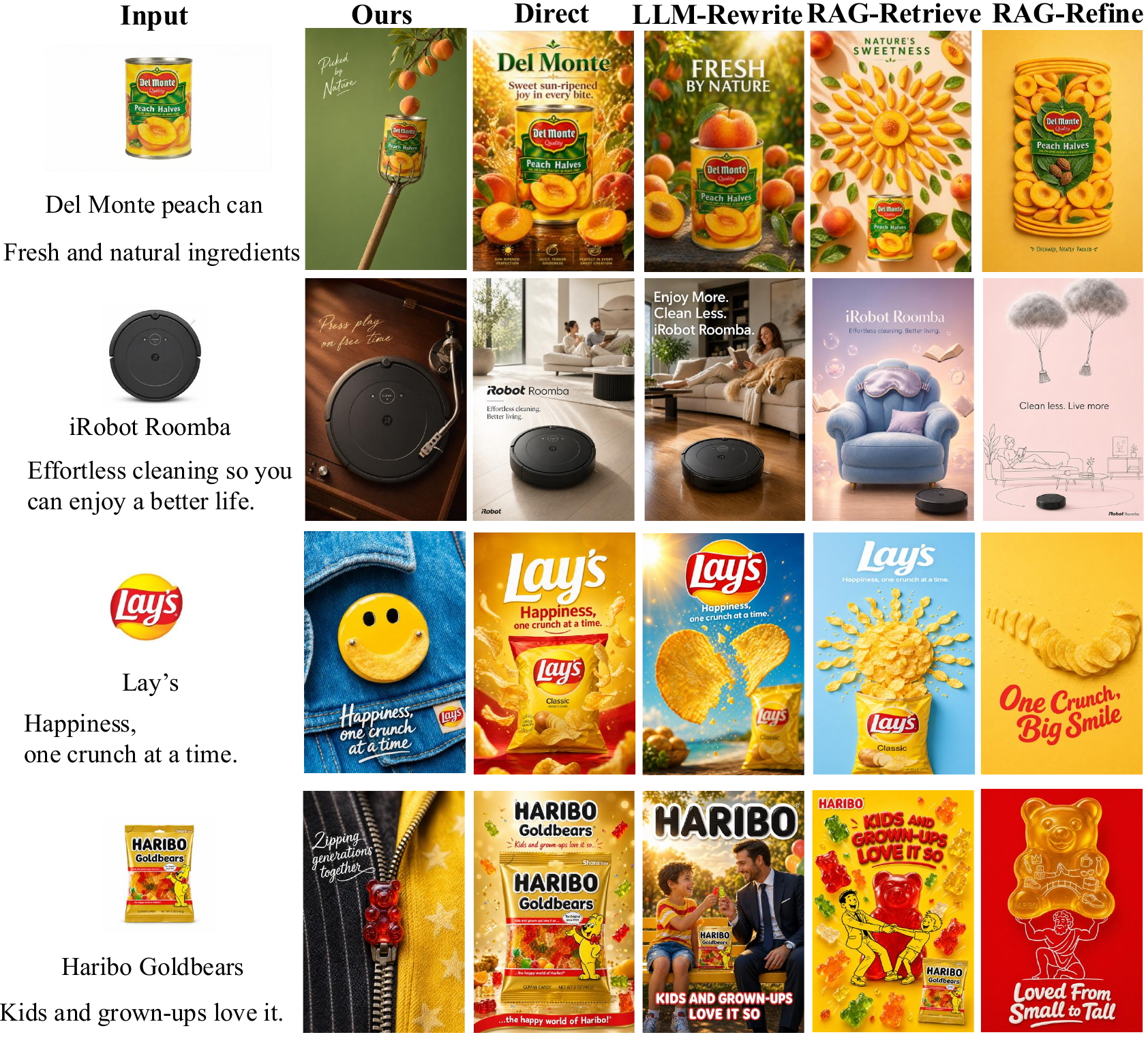}
  \caption{Qualitative evaluation results.}
  \label{fig:baseline}
\end{figure}

As shown in Fig.~\ref{fig:baseline}, the baselines tend to either render the brief literally or add visual polish without forming a cross-domain mapping. In contrast, \sysname more consistently translates the positioning into a concrete cross-domain visual concept while maintaining product visibility: the Del Monte can appears as fruit freshly ``picked by nature'' among peach branches, returning packaged product to its origin environment; the Roomba becomes a vinyl record, recasting effortless cleaning as pressing play on free time; a Lay's chip turns into a smiling button on a denim jacket by form analogy; and a gummy bear serves as a zipper pull that zips generations together. In each case, the product assumes a role in the target domain rather than merely appearing in a themed scene. 

\section{Conclusion}
\label{sec:conclusion}

This paper introduced \sysname, a skill-guided multi-agent framework for creative advertising image generation. We distill recurring cross-domain \emph{creative bridges} from high-quality ads into reusable \emph{creative skills}. \sysname uses these skills for strategy routing, multi-candidate ideation and selection, and VLM-guided refinement.
Future work can explore more explicit controls over creativity, alignment, and brand preference in advertising workflows.

\end{sloppypar}

\begin{acks}
This work was supported in part by the National Natural Science Foundation of China under no. 62572458, in part by the Beijing Natural Science Foundation under no. 26QY0538, and in part by the National Science and Technology Council, Taiwan under no. 113-2221-E006-161-MY3.
\end{acks}

\bibliographystyle{ACM-Reference-Format}
\bibliography{AdMuse}

\end{document}


\title{Supplementary Materials of AdMuse: Skill-Guided Creative Advertising Image Generation}

\settopmatter{printacmref=false}
\maketitle

\renewcommand{\thesection}{S\arabic{section}}
\renewcommand{\thetable}{S\arabic{table}}
\renewcommand{\thefigure}{S\arabic{figure}}
\setcounter{section}{0}
\setcounter{table}{0}
\setcounter{figure}{0}

\begin{sloppypar}

%

\newlist{schema}{itemize}{1}
\setlist[schema]{
  label={\textendash},
  leftmargin=1.4em, labelsep=0.5em,
  itemsep=2pt, parsep=0pt, topsep=3pt
}

\definecolor{ideaC}{HTML}{2E7CD6}     
\definecolor{judgeC}{HTML}{6B5BB5}    
\definecolor{promptC}{HTML}{D97706}   
\definecolor{criticC}{HTML}{B4322A}   
\definecolor{captionC}{HTML}{0E7C66}  
\definecolor{evalC}{HTML}{4B5563}     
\definecolor{baseDC}{HTML}{8B6F4F}    
\definecolor{baseLC}{HTML}{5F7159}    
\definecolor{baseRC}{HTML}{6F5180}    
\definecolor{baseFC}{HTML}{1F6F78}    

\newtcolorbox{promptcard}[2][]{
  enhanced jigsaw, breakable,
  colback=#2!4,
  colframe=#2!70,
  coltitle=white,
  colbacktitle=#2!70,
  fonttitle=\bfseries\small\sffamily,
  attach boxed title to top left={xshift=10pt, yshift=-10pt},
  boxed title style={
    colframe=#2!70, colback=#2!70,
    sharp corners, arc=2pt, boxrule=0pt,
    left=6pt, right=6pt, top=2pt, bottom=2pt
  },
  arc=4pt, boxrule=0.6pt,
  left=8pt, right=8pt, top=14pt, bottom=8pt,
  fontupper=\small,
  before skip=8pt, after skip=10pt,
  #1
}

\newtcolorbox{prompttext}{
  enhanced, breakable,
  colback=gray!6, colframe=gray!45,
  boxrule=0.3pt, arc=2pt,
  left=5pt, right=5pt, top=4pt, bottom=4pt,
  fontupper=\ttfamily\scriptsize,
  before skip=4pt, after skip=4pt
}

\section{Dataset and Skill Distillation}
\label{sec:supp_dataset_distillation}

\subsection{Dataset and Distillation Corpus}
\label{sec:supp_dataset}

\begin{figure}[t]
  \centering
  \includegraphics[width=1\linewidth]{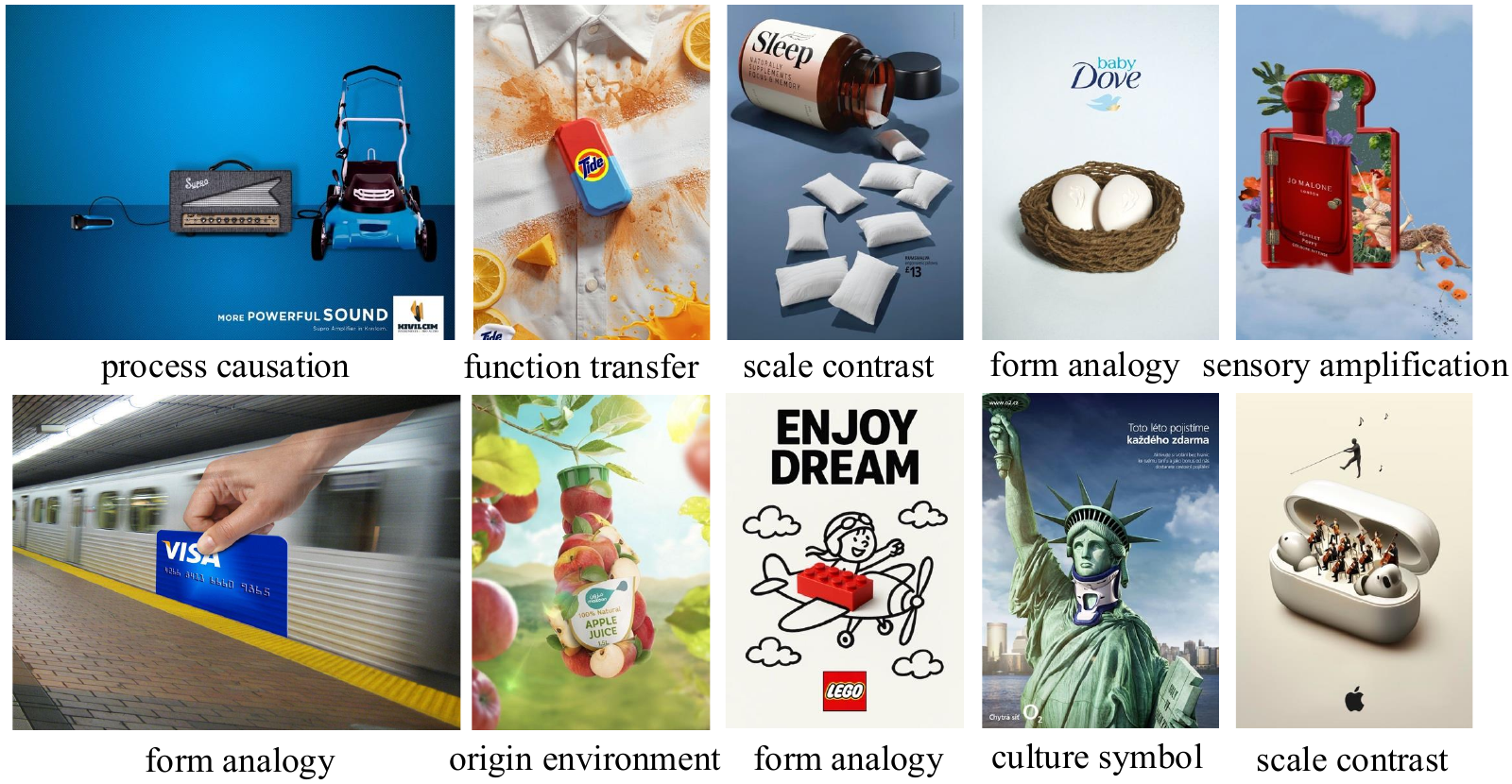}
  \caption{Some creative advertising images and their corresponding bridge types in our dataset.}
  \label{fig:dataset}
\end{figure}

Figure~\ref{fig:dataset} shows representative advertisements from the distillation corpus and their annotated bridge types.
In total, we collected 1,129 high-quality creative advertisements from public creative platforms such as Xiaohongshu and Pinterest, covering a wide range of industries, formats, and bridge types.

\subsection{Caption-Generation Prompt}
\label{sec:supp_caption_prompt}

The $1{,}129$ advertisements are captioned by a single Gemini-3.1-Pro VLM call per image.
The prompt elicits a three-layer JSON record---(L1)~product identity, (L2)~associative space, (L3)~creative bridge---and commits the image to exactly one of eight predefined bridge-type labels.
The \texttt{visual\_structure} and \texttt{image\_prompt} fields are reused both by skill distillation and by the RAG baselines.

\begin{promptcard}[title={Caption Generator -- Three-Layer Ad Analysis}]{captionC}
\emph{Role.} A senior creative director analysing a high-quality advertising image to extract its \emph{creative mechanism}.\\
\emph{Layer 1 -- Product identity.} Describe what the product \emph{is}: \texttt{physical\_form} (shape, colour, material, texture, size), \texttt{function} (what it does, how users interact with it), \texttt{ingredient\_origin} (what it is made from), \texttt{brand\_associations} (logo symbolism, cultural connotations), and a single-sentence \texttt{communicated\_positioning}.\\
\emph{Layer 2 -- Associative space.} Describe what the product \emph{reminds the viewer of}: \texttt{visual\_resemblance} (objects sharing a similar look), \texttt{semantic\_links} (conceptual links through function, ingredients, brand meaning, or culture), and \texttt{contextual\_world} (environment suggested by the association).\\
\emph{Layer 3 -- Creative bridge.} Describe \emph{how} the ad exploits the association: an \texttt{associative\_chain} written as ``A $\to$ B $\to$ C $\to$ D,'' exactly one \texttt{bridge\_type} drawn from the taxonomy, and a one-sentence \texttt{why\_it\_works}.\\
\emph{Execution and reasoning fields.} The model additionally writes (i)~a single-sentence \texttt{visual\_structure} describing composition, lighting, palette, and subject placement; (ii)~a 3--4 sentence \texttt{creative\_reasoning} explaining why the visual expresses the positioning; and (iii)~an \texttt{image\_prompt} of $\le\!150$ words, framed as a text-to-image instruction that could recreate the ad---containing only visible content, ending with technical quality tags, and never mentioning brand names or abstract strategy.

\emph{Return schema:}
\begin{schema}
\item \texttt{product\_subject} --- short noun phrase identifying the product.
\item \texttt{communicated\_positioning} --- one-sentence positioning takeaway.
\item \texttt{product\_identity} --- object with \texttt{physical\_form}, \texttt{function}, \texttt{ingredient\_origin}, \texttt{brand\_associations}; each a list of short phrases.
\item \texttt{associative\_space} --- object with \texttt{visual\_resemblance}, \texttt{semantic\_links}, \texttt{contextual\_world}; each a list of short phrases.
\item \texttt{creative\_bridge} --- object with \texttt{associative\_chain}, \texttt{bridge\_type}, \texttt{why\_it\_works}.
\item \texttt{execution} --- object with \texttt{visual\_structure} (single sentence).
\item \texttt{creative\_reasoning} --- 3--4 sentence justification.
\item \texttt{image\_prompt} --- $\le\!150$ word visible-content-only text-to-image instruction.
\end{schema}
\end{promptcard}
\subsection{Running Skill Example: S018}
\label{sec:supp_skill_example}

The main-paper pipeline overview uses S018, \emph{Product Substitution In Scene}, as its winning skill. It belongs to the \texttt{form\_analogy} bridge type and encodes the following strategy: substitute a product, its packaging, or one of its pieces for a geometrically similar object in a recognizable scene, so that the substituted object communicates the product's functional or emotional benefit.

\paragraph{Skill representation.}
Each distilled skill stores (i) a strategy description, (ii) applicability cues and anti-triggers organized by physical form, functional role, ingredient origin, and brand theme, (iii) a parameterized associative-chain template, and (iv) a playbook for constructing the visual concept. S018 uses the template
\begin{equation}
\begin{aligned}
\{\mathrm{product\ shape}\}
&\rightarrow \{\mathrm{metaphorical\ object}\}\\
&\rightarrow \{\mathrm{core\ benefit}\}
\rightarrow \mathrm{product\ substitution\ in\ scene}.
\end{aligned}
\label{eq:s018-template}
\end{equation}
The skill is applicable when the product has a distinctive, simple silhouette and a familiar object with a matching shape can express the intended benefit. It should be avoided when the resemblance is weak, the product form is irregular or mutable, or the substitution requires a cluttered scene that obscures the visual analogy.

\paragraph{Instantiation procedure.}
The playbook instructs the IdeaAgent to: (1) isolate the product's primary silhouette from a flat or orthographic view; (2) identify familiar objects that both match this silhouette and embody the positioning; (3) establish a minimal scene that makes the target object recognizable; and (4) replace that object with the product while matching its scale, orientation, and alignment. A stylistic contrast---typically a photorealistic product against line art or simplified contextual elements---keeps the product visually dominant. In the main-paper pipeline example, these instructions lead the agent to treat the open Campus notebook as a stage on which its margin doodles perform, communicating the positioning that the notes become the main characters.

\section{Pipeline Details}
\label{sec:supp_pipeline}

\subsection{Skill Routing Details}
\label{sec:supp_routing}

For a brief $x$ and skill $s$, the SkillRouter computes
\begin{equation}
\begin{aligned}
\mathrm{score}(s,x)={}&w_{\mathrm{trig}}\phi_{\mathrm{trig}}
+w_{\mathrm{desc}}\phi_{\mathrm{desc}}
+w_{\mathrm{Elo}}\hat{R}_{s}\\
&+w_{\mathrm{bucket}}\hat{R}_{s,c}
+w_{\mathrm{expl}}U_s+\delta_{\mathrm{anti}},
\end{aligned}
\label{eq:supp-router}
\end{equation}
where $\phi_{\mathrm{trig}}$ and $\phi_{\mathrm{desc}}$ are cosine similarities between the brief embedding and the skill's triggers and description; $\hat{R}_{s}$ and $\hat{R}_{s,c}$ are normalized global and category-specific skill evidence; $U_s$ favors less frequently tested skills; and $\delta_{\mathrm{anti}}\!\leq\!0$ penalizes overlap with anti-triggers. A small user-controlled Gumbel perturbation introduces ranking stochasticity. Starting from the highest-ranked candidate, the router greedily selects three skills while penalizing bridge types already represented in the selected set.

\subsection{Visual Refinement Details}
\label{sec:supp_realization}

The PromptAgent first converts the winning concept $y^*$ into an image prompt $u_0$. At round $r$, the renderer produces $\mathbf{I}_r$, and the CriticAgent evaluates concept fidelity, product and brand preservation, composition, and rendering quality. If rejected, a VLM compresses the prompt, rendered image, and critique into an experience record of at most 100 words. This record is appended to the cumulative memory $M_r$, allowing the next prompt to preserve successful elements while avoiding previously observed failures:
\begin{equation}
\mathbf{I}_r=\mathcal{G}_{\mathrm{img}}(u_r),\qquad
u_{r+1}=\mathcal{G}_{\mathrm{prompt}}\!\left(u_r,\mathcal{G}_{\mathrm{critic}}(\mathbf{I}_r,y^*),M_r\right).
\label{eq:supp-realization}
\end{equation}
The final image is returned when approved or when $R_{\max}=3$ rounds are reached.

\subsection{Agent Prompts}
\label{sec:supp_agent_prompts}

We summarise each of the four pipeline agents below as a quadruple of \emph{role} / \emph{mandate} / \emph{inputs} / \emph{procedure}, followed by the JSON return schema.
The \textsc{IdeaAgent} and \textsc{PromptAgent} use GPT-5.5, while the \textsc{JudgeAgent} and \textsc{CriticAgent} use Gemini-3.1-Pro.
Placeholders such as \texttt{\{positioning\}} are filled at runtime from the brief.
The full verbatim prompt text is shipped in the accompanying supplementary release.

\begin{promptcard}[title={IdeaAgent (S1) -- Skill-Conditioned Brainstorming}]{ideaC}
\emph{Role.} A senior advertising creative director generating one idea per routed skill.\\
\emph{Mandate.} The positioning is the \emph{destination} and the routed skill is the \emph{route}; an idea creative but drifting from the positioning is declared failed.\\
\emph{Inputs.} Brief fields (\texttt{product\_name}, \texttt{positioning}, \texttt{task\_type}, \texttt{target\_audience}, \texttt{negative\_constraints}) and the selected skill's \texttt{\{skill\_block\}} playbook.\\
\emph{Procedure.} (i)~decode the specific emotion or truth captured by the positioning; (ii)~apply the skill as a vehicle; (iii)~concretise visual subjects, composition, lighting, colour, and mood; (iv)~self-check whether a stranger would immediately feel the positioning.
Hard rules forbid literal product shots and advertising clich\'es and require brand-safe content; an optional tagline is capped at $8$ words.

\emph{Return schema:}
\begin{schema}
\item \texttt{positioning\_decode} --- decoded emotional truth of the positioning.
\item \texttt{metaphor\_core} --- one-line statement of the central metaphor.
\item \texttt{visual\_concept} --- paragraph describing the intended visual.
\item \texttt{emotional\_tone} --- short tone label.
\item \texttt{tagline} --- $\le\!8$ words, or empty string.
\item \texttt{tagline\_hint} --- guidance for typography placement.
\item \texttt{reasoning} --- explanation of how the skill serves the positioning.
\end{schema}
\end{promptcard}

\begin{promptcard}[title={JudgeAgent (S2) -- Pairwise Winner Selection}]{judgeC}
\emph{Role.} A senior creative director selecting the best of $n$ candidate concepts \emph{relatively}, without absolute scoring.\\
\emph{Inputs.} The brief plus a concatenated \texttt{\{candidates\_block\}} listing the $n$ IdeaAgent outputs.\\
\emph{Procedure.} Compare candidates in the priority order (1)~positioning alignment, (2)~creativity / surprise, (3)~visual feasibility for a single still image, (4)~skill-execution quality as tiebreaker; pick exactly one winner.

\emph{Return schema:}
\begin{schema}
\item \texttt{winner\_index} --- $0$-based index of the chosen candidate.
\item \texttt{reasoning} --- 2--3 sentences justifying the relative choice.
\end{schema}
\end{promptcard}

\begin{promptcard}[title={PromptAgent (S3) -- Idea-to-Prompt Translation}]{promptC}
\emph{Role.} A specialist that converts the winning idea into a single image-generation prompt of $<\!300$ words.\\
\emph{Inputs.} \texttt{product\_name}, \texttt{metaphor\_core}, \texttt{visual\_concept}, \texttt{emotional\_tone}, \texttt{task\_type}, and \texttt{negative\_constraints}; the with-text variant additionally consumes a text-design block (content, placement, font personality, stroke dynamics, palette with hex codes) and embeds the typography as an integral visual element.\\
\emph{Procedure.} The prompt must specify composition (foreground/background placement), lighting style, palette keywords, and end with technical quality tags (e.g.\ ``photorealistic, 8K, award-winning commercial photography''); the metaphor itself is \emph{not} written into the prompt---only what is visible.\\
\emph{Product-anchor rule.} If the visual already contains the product form, do not add a second copy; only when the product/brand is completely absent should a small subtle integrated instance be placed in one corner.\\
\emph{Revision variant.} On rounds $r{>}1$ the agent additionally receives the full optimisation trajectory (prior prompts, rendered images, critic feedback, compressed experience records) and is instructed to (i)~avoid strategies already flagged as failed, (ii)~preserve elements the critic marked as effective, and (iii)~address only the issues raised in the latest critique.\\
\emph{Output.} A single plain-text prompt; no JSON, no preamble.
\end{promptcard}

\begin{promptcard}[title={CriticAgent (S4) -- VLM Image Critique}]{criticC}
\emph{Role.} A senior art director deciding whether the rendered image is ready for publication.\\
\emph{Inputs.} The brief, the intended creative idea (\texttt{metaphor\_core}, \texttt{visual\_concept}, \texttt{emotional\_tone}), and the rendered image attached as a VLM input.\\
\emph{Procedure.} Evaluate the image \emph{independently} on three dimensions: (D1)~technical quality---hard gate, fails only on obvious artefacts or severe compositional problems; (D2)~creative effectiveness---main gate, passes if the core idea is recognisable and the positioning is communicated even if execution is imperfect; (D3)~visual attraction---soft gate, requires both aesthetic appeal and absence of brand-inappropriate content.
The boolean \texttt{approved} is the conjunction of the three pass flags.

\emph{Return schema:}
\begin{schema}
\item \texttt{quality\_pass} --- boolean; D1 outcome.
\item \texttt{quality\_note} --- short artefact / composition note.
\item \texttt{effectiveness\_pass} --- boolean; D2 outcome.
\item \texttt{effectiveness\_note} --- short concept / positioning note.
\item \texttt{aesthetic\_pass} --- boolean; D3 outcome.
\item \texttt{aesthetic\_note} --- short aesthetic / safety note.
\item \texttt{approved} --- conjunction of the three pass flags.
\item \texttt{overall\_feedback} --- revision instructions per failed dimension.
\item \texttt{preserve\_note} --- elements that are working and should be preserved.
\item \texttt{reasoning} --- independent analysis of each dimension.
\end{schema}
\end{promptcard}

\subsection{Hyperparameters}
\label{sec:supp_hparams}

Table~\ref{tab:hparams} consolidates the loop, router, and agent constants used by all reported runs unless explicitly varied.

\begin{table}[h]
\centering
\small
\caption{Pipeline, router, and agent hyperparameters.}
\label{tab:hparams}
\begin{tabular}{lll}
\toprule
Group & Parameter & Value \\
\midrule
\multirow{2}{*}{Pipeline loops}
  & $R_{\max}$ (image refinement)         & 3 \\
  & Idea candidates per brief             & 3 \\
\midrule
\multirow{6}{*}{Router}
  & $w_{\mathrm{trig}}, w_{\mathrm{desc}}$        & 0.20, 0.40 \\
  & $w_{\mathrm{elo}}, w_{\mathrm{bkt}}$          & 0.15, 0.10 \\
  & $w_{\mathrm{expl}}$ (UCB bonus)               & 0.15 \\
  & MMR $\lambda$                                 & 0.70 \\
  & $k$ skills per run                            & 3 \\
  & Embedding model                               & Qwen3-Embedding-0.6B \\
\midrule
\multirow{6}{*}{Agent temperature}
  & IdeaAgent brainstorm / refine         & 0.85 / 0.70 \\
  & PromptAgent craft / revise            & 0.60 / 0.50 \\
  & JudgeAgent                            & 0.15 \\
  & CriticAgent                           & 0.15 \\
  & Memory compression                    & 0.20 \\
  & Category inference                    & 0.10 \\
\bottomrule
\end{tabular}
\end{table}

\section{Experimental Protocol}
\label{sec:supp_protocol}

\subsection{Benchmark Composition}
\label{sec:supp_benchmark}

\begin{figure}[H]
  \centering
  \includegraphics[width=0.8\linewidth]{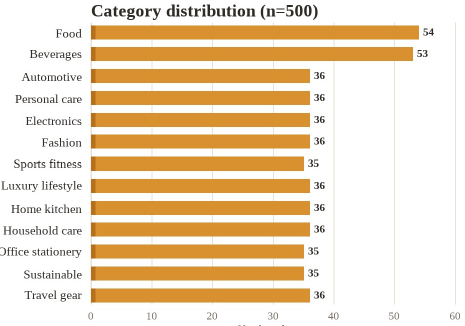}
  \caption{Category distribution of the 500 benchmark briefs.}
  \label{fig:supp_category_distribution}
\end{figure}

The benchmark contains $500$ advertising briefs drawn from a held-out product pool covering $13$ commercial categories: \textbf{\emph{food}, \emph{beverages}, \emph{automotive}, \emph{personal\_care}, \emph{electronics}, \emph{fashion\_apparel}, \emph{sports\_fitness}, \emph{luxury\_lifestyle}, \emph{home\_kitchen}, \emph{household\_care}, \emph{office\_stationery}, \emph{sustainable\_products}, and \emph{travel\_gear}}, illustrated in Fig~\ref{fig:supp_category_distribution}.
Across the $500$ briefs there are  $420$ unique product reference images; a subset of $59$ products carries more than one positioning, enabling controlled comparison of how the same product is interpreted under different brand messages.
The benchmark briefs are excluded from the $1{,}129$-ad distillation corpus so that no skill has been trained on a benchmark item.

\subsection{Baseline Implementations}
\label{sec:supp_baselines}

We compare AdMuse against four baselines spanning direct prompting, LLM rewriting, retrieval augmentation, and retrieval-guided refinement.
\textsc{Direct} sends the brief straight to the image renderer; \textsc{LLM-Rewrite} uses one LLM call to expand the brief into a detailed image-generation prompt; \textsc{RAG-Retrieve} retrieves a single high-level visual sketch from the captioned reference corpus and uses it as inspiration; \textsc{RAG-Refine} retrieves three structured records and runs the same idea--judge--prompt--critic loop as AdMuse, but without skill routing or skill templates.
All methods share the same image-rendering backend used by AdMuse, so quality differences are attributable to the information and reasoning given to the renderer rather than the renderer itself.

\begin{promptcard}[title={Baseline 1 -- Direct}]{baseDC}
\emph{Pipeline.} The brief fields are concatenated into a single text-to-image instruction; no LLM is interposed between brief and renderer.
This baseline isolates what the image model can do unaided.\\
\emph{Inputs.} \texttt{product\_name}, \texttt{positioning}, \texttt{task\_type}, \texttt{target\_audience}.\\
\emph{Calls.} 1 image-generation call; 0 LLM calls.

\emph{Prompt template (sent directly to the image backend):}
\begin{prompttext}
A commercial advertising image for \{product\_name\}.\\
Positioning: \{positioning\}.\\
Task type: \{task\_type\}; target audience: \{target\_audience\}.\\
Photorealistic, 8K, award-winning commercial photography.
\end{prompttext}
\end{promptcard}

\begin{promptcard}[title={Baseline 2 -- LLM-Rewrite}]{baseLC}
\emph{Pipeline.} A single LLM call expands the brief into a detailed image-generation prompt, which is then rendered by the same backend.
This isolates the contribution of LLM-mediated prompt engineering without any external knowledge.\\
\emph{Inputs.} Same brief fields as Baseline~1.\\
\emph{Calls.} 1 LLM call (prompt expansion) + 1 image-generation call.

\emph{LLM prompt template:}
\begin{prompttext}
[SYSTEM]\\
You are a senior advertising creative director. Given the brief below, write a single image-generation prompt of $<\!300$ words that visually communicates the positioning. Specify composition (foreground / background placement), lighting style, colour palette, and end with technical quality tags. Avoid literal product shots; prefer metaphor or symbolic imagery where appropriate. Return ONLY the prompt text, no preamble.\\[3pt]
[USER]\\
Product / Brand: \{product\_name\}\\
Positioning:     \{positioning\}\\
Task type:       \{task\_type\}\\
Target audience: \{target\_audience\}\\
Negative constraints: \{negative\_constraints\}
\end{prompttext}
\end{promptcard}

\begin{promptcard}[title={Baseline 3 -- RAG-Retrieve}]{baseRC}
\emph{Pipeline.} A high-level visual sketch is retrieved from the same captioned corpus used for skill distillation and shown to the LLM as inspiration; the LLM then writes one image-generation prompt conditioned on both the brief and the retrieved sketch.
The retrieved field is the abstract \texttt{visual\_structure}, not the ready-to-render \texttt{image\_prompt}, isolating the contribution of \emph{retrieval-as-inspiration} from verbatim copying.\\
\emph{Inputs.} Brief fields plus the top-$1$ retrieved \texttt{visual\_structure}.\\
\emph{Calls.} 1 FAISS retrieval + 1 LLM call + 1 image-generation call.

\emph{Retrieval details.}
\begin{itemize}[nosep,leftmargin=*,topsep=2pt]
  \item \textbf{Corpus.} $1{,}129$ entries from \texttt{ref\_txt/}; indexed key is \texttt{execution.visual\_structure} (a 2--3 sentence sketch of composition, lighting, palette, and subject placement).
  \item \textbf{Encoder.} \texttt{Qwen3-Embedding-0.6B} (local sentence-transformer).
  \item \textbf{Index.} \texttt{faiss.IndexFlatIP} over L2-normalised embeddings; cosine similarity reduces to inner product.
  \item \textbf{Query.} concatenation of \texttt{product\_name}, \texttt{positioning}, \texttt{task\_type}, and \texttt{target\_audience}, joined by ``\,|\,''; missing fields are dropped.
  \item \textbf{Retrieval.} top-$1$ ($k{=}1$); only \texttt{visual\_structure} is shown to the LLM, with an explicit instruction not to copy literally.
\end{itemize}

\emph{LLM prompt template:}
\begin{prompttext}
[SYSTEM]\\
You are a senior advertising creative director. Given the brief and a retrieved visual sketch of a related high-quality ad, write a single image-generation prompt of $<\!300$ words that visually communicates the positioning. Use the retrieved sketch as INSPIRATION ONLY -- do not copy its subjects literally. Specify composition, lighting, colour palette, and end with technical quality tags. Return ONLY the prompt text, no preamble.\\[3pt]
[USER]\\
Product / Brand: \{product\_name\}\\
Positioning:     \{positioning\}\\
Task type:       \{task\_type\}\\
Target audience: \{target\_audience\}\\
Negative constraints: \{negative\_constraints\}\\[2pt]
Retrieved visual sketch (inspiration only, do not copy):\\
\{visual\_structure\}
\end{prompttext}
\end{promptcard}

\begin{promptcard}[title={Baseline 4 -- RAG-Refine}]{baseFC}
\emph{Pipeline.} The top-$3$ records are retrieved from the same captioned corpus and passed to an agentic refinement loop that mirrors AdMuse's ideation and realization stages.
Unlike AdMuse, this baseline receives retrieved examples rather than distilled skills, and it performs no skill routing or template binding.\\
\emph{Inputs.} Brief fields plus the top-$3$ retrieved records, including product identity, associative chain, rationale, and \texttt{visual\_structure}.\\
\emph{Calls.} 1 FAISS retrieval + 3 IdeaAgent calls + 1 JudgeAgent call + up to $R_{\max}$ PromptAgent / image-generation / CriticAgent rounds.

\emph{Retrieval details.}
\begin{itemize}[nosep,leftmargin=*,topsep=2pt]
  \item \textbf{Corpus and encoder.} Same $1{,}129$ captioned records and \texttt{Qwen3-Embedding-0.6B} index as RAG-Retrieve.
  \item \textbf{Retrieval.} top-$3$ ($k{=}3$); each retrieved record is treated as a high-level creative reference rather than a renderable prompt.
  \item \textbf{Control.} The downstream idea--judge--prompt--critic loop is kept identical to AdMuse, isolating whether retrieved exemplars can replace the skill library.
\end{itemize}
\end{promptcard}

\subsection{Evaluation Rubric and Judge Prompt}
\label{sec:supp_rubrics}

We evaluate each generated advertisement along two complementary dimensions: creativity measures the originality and visual impact of the concept, while alignment measures how clearly the image communicates the requested product positioning. Two independent out-of-loop VLM judges, Claude Sonnet 4.6 and Qwen3.7-Plus, assign both scores on explicit ten-point scales using the same prompt and level-specific criteria in Tables~\ref{tab:cs-rubric} and~\ref{tab:as-rubric}.

\begin{promptcard}[title={Independent Evaluators -- Out-of-Loop Judges}]{evalC}
\emph{Role.} An \emph{external} VLM judge---separate from the in-loop Gemini-3.1-Pro \textsc{JudgeAgent} and \textsc{CriticAgent}---scoring the final rendered ad against the brief. Claude Sonnet 4.6 and Qwen3.7-Plus are used only in evaluation and do not participate in the generation pipeline.\\
\emph{Inputs.} The generated image, the original product reference image(s), and the brief.\\
\emph{Procedure.} Return two integer scores in $[1,10]$ with rationale fields.
The judge is instructed to (i)~anchor each score against the explicit rubric level rather than gut feeling; (ii)~use the full $1$--$10$ range so that ``$5$'' denotes \emph{mediocre, barely functional} rather than ``average''; (iii)~score the generated image alone---no credit is transferred from the reference image; and (iv)~penalise plain product display with no visual effort to communicate the positioning ($\le\!4$ on alignment), since product visibility alone does not constitute positioning communication.
Tables~\ref{tab:cs-rubric} and~\ref{tab:as-rubric} reproduce the two ten-level rubrics.

\emph{Return schema:}
\begin{schema}
\item \texttt{creativity\_rationale} --- rubric-level reference and visible-evidence summary.
\item \texttt{creativity\_score} --- integer in $[1,10]$.
\item \texttt{alignment\_rationale} --- rubric-level reference and visible-evidence summary.
\item \texttt{alignment\_score} --- integer in $[1,10]$.
\end{schema}
\end{promptcard}

\begin{table}[h]
\centering
\small
\caption{Creativity-score rubric: \emph{creative impact} of the generated ad.}
\label{tab:cs-rubric}
\begin{tabular}{rp{6.2cm}}
\toprule
Score & Descriptor \\
\midrule
10 & Extraordinary; award-worthy concept with flawless execution; immediate aha moment via unexpected metaphor, paradox, or conceptual leap. \\
 9 & Outstanding; strong original concept clearly visible; memorable metaphor. \\
 8 & Very good; clear creative idea beyond product display; recognisable device that feels fresh. \\
 7 & Good; discernible concept above generic display, but metaphor is somewhat expected or execution under-realises it. \\
 6 & Above average; concept attempted but too subtle, conventional, or partly obscured by execution problems. \\
 5 & Mediocre; minor decorative elements beyond plain display, no clear creative hook. \\
 4 & Below average; styled product shot, no creative concept, metaphor, or visual surprise. \\
 3 & Weak; plain product display with minimal styling, or a concept that fails to communicate. \\
 2 & Poor; incoherent visuals or severely flawed attempt where creative intent is unrecognisable. \\
 1 & Failed; incoherent, off-topic, or no advertising intent. \\
\bottomrule
\end{tabular}
\end{table}

\begin{table}[h]
\centering
\small
\caption{Alignment-score rubric: \emph{communication alignment} between image and positioning.}
\label{tab:as-rubric}
\begin{tabular}{rp{6.2cm}}
\toprule
Score & Descriptor \\
\midrule
10 & Positioning unambiguously communicated through visual means; product/brand unmistakable; creative choices directly reinforce positioning. \\
 9 & Positioning clearly communicated; product prominent; only minor concept-positioning gap. \\
 8 & Positioning communicated effectively through visual means; a different positioning would require different visual choices. \\
 7 & Positioning visually present but requires interpretation; visual--positioning link not tight. \\
 6 & Positioning partially communicated through subject or scene; visuals could serve a different positioning with minor changes. \\
 5 & Positioning vague or generic; subject and composition make no specific attempt beyond showing the product (overlaid slogan text does not count as visual communication). \\
 4 & Product display with no positioning conveyed visually; positioning barely recognisable from the image alone. \\
 3 & Little to no visual connection to the requested positioning. \\
 2 & Visual contradicts the positioning, or product/brand absent. \\
 1 & No connection whatsoever; wrong product, message, or image. \\
\bottomrule
\end{tabular}
\end{table}

\end{sloppypar}

\bibliographystyle{ACM-Reference-Format}
\bibliography{AdMuse}